\RequirePackage{fix-cm}
\documentclass[10pt,letterpaper]{article}
\usepackage{spconf,amsmath,amssymb,graphicx,booktabs,microtype,cite,url}
\microtypesetup{protrusion=false}
\graphicspath{{figures/}}

\newcommand{\DC}{D_{\mathrm C}}
\newcommand{\DS}{D_{\mathrm S}}
\title{Evaluator-Dependent Patient-Adaptive ECG Lead-Channel Allocation}
\name{Xiaoyang Li \qquad Zeyan Tao}
\address{College of Medicine and Biological Information Engineering\\
Northeastern University, Shenyang, China\\
\{20246389@stu.neu.edu.cn, taozy@mails.neu.edu.cn\}\\
ORCID: 0009-0002-4863-5761 (X. Li); 0009-0000-3149-9105 (Z. Tao)}
\DeclareMathSizes{10}{10}{9.2}{9.2}
\DeclareMathSizes{9}{9.2}{9.2}{9.2}

\makeatletter
\newcommand{\figcaption}{\def\@captype{figure}\caption}
\makeatother
\let\originalbibliography\thebibliography
\renewcommand{\thebibliography}[1]{\originalbibliography{#1}\setlength{\itemsep}{2pt}\setlength{\parsep}{0pt}}
\begin{document}
\pdfinfo{/Title (Evaluator-Dependent Patient-Adaptive ECG Lead-Channel Allocation) /Author (Xiaoyang Li; Zeyan Tao)}
\maketitle
\begin{abstract}
The diagnostic value of an ECG channel depends on the model that interprets it.
We test whether adaptive acquisition retains its advantage over fixed protocols
when that model changes. Two policies developed with a logistic evaluator
are frozen and assessed with a masked waveform ResNet1D, using exhaustive,
metric-matched fixed comparators. On PTB-XL, replacing the evaluator reverses
the mean adaptive advantage in negative log-likelihood and Brier score
across the tested budgets, while calibration responds less uniformly.
The shift persists with common fixed references and broader training-mask
exposure. Training a policy against the stronger evaluator partly recovers
the lost advantage. These retrospective exploratory
results show that acquisition quality cannot be assessed independently of the
downstream evaluator and motivate validating adaptive policies with the model
intended for diagnosis.
\end{abstract}

\section{Introduction}
The standard 12-lead electrocardiogram (ECG) captures complementary views of
cardiac electrical activity~\cite{kligfield2007recommendations}. Reduced-lead
systems often use a population-wide subset~\cite{nelwan2004reconstruction,reyna2021will,lai2021optimal},
although the diagnostic value of an additional channel may vary across
patients. A patient-adaptive policy instead selects channels sequentially
from the partially observed recording. Here, the budget $k$ counts accessed
lead channels, not electrodes or physical acquisition actions.

Channel utility also depends on the diagnostic evaluator. A policy developed
with a lightweight model may exploit features that a replacement model uses
differently. Better prediction from complete recordings therefore does not
guarantee that the same adaptive subsets remain preferable to a fixed protocol.
We test this distinction by freezing two policies trained with a controlled
logistic evaluator and scoring their unchanged subsets with a masked waveform
ResNet1D. Exhaustive search provides a metric-matched fixed comparator for
each evaluator.

Our contribution is an empirical test of evaluator dependence in adaptive
ECG allocation. We measure the change in the adaptive-minus-fixed contrast,
then examine common fixed references, broader training-mask exposure, and
evaluator-aligned policy training. Figure~\ref{fig:architecture} summarizes the
design. The held-out analysis is exploratory, and the additional sensitivity
analyses are post hoc.

\section{Related Work}
Reduced-lead ECG research includes fixed configurations, cohort-level selection,
and reconstruction~\cite{nelwan2004reconstruction,reyna2021will,lai2021optimal}.
Lead-agnostic pretraining and arbitrary-lead reconstruction handle missing
channels~\cite{oh2022leadagnostic,chen2024mcma}, whereas lead switching addresses
sequential observation~\cite{iwata2024switching}. Waveform convolutional networks
provide strong diagnostic evaluators~\cite{ribeiro2020automatic,strodthoff2021deep}.

Active feature acquisition selects an unavailable feature from the current
state~\cite{shim2018pay,janisch2019costly}, using conditional models or information
gain~\cite{ma2019eddi,li2020acflow,gong2019icebreaker,covert2023dynamic}.
Task-driven sensing similarly couples measurements to an inference
objective~\cite{duarte2013taskdriven,chakrabarti2016sensor}. We study a specific
consequence of this dependence: whether an adaptive ECG advantage survives
replacement of the evaluator. We assess probability quality separately from
ranking performance~\cite{vancalster2019calibration}.

\section{Method}\label{sec:method}
\subsection{Exact-budget acquisition}
\begin{figure}[p]
\centering
\includegraphics[width=\textwidth]{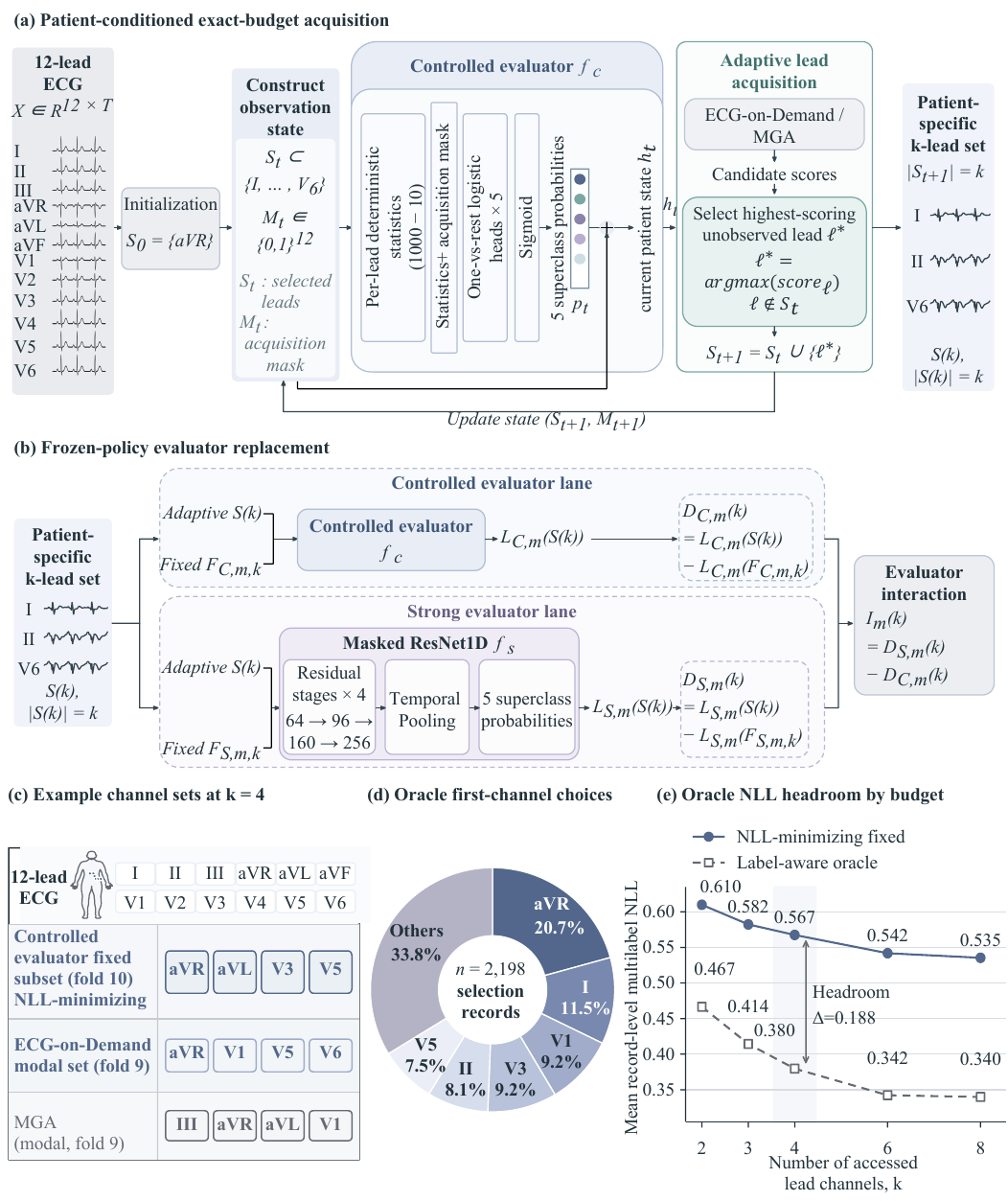}
\caption{(a) Adaptive acquisition and (b) frozen-policy evaluator replacement.
Descriptive pre-holdout evidence: (c) channel sets, (d) oracle first-channel
choices, and (e) NLL headroom. Oracles use labels and are non-deployable.
The depicted aVR seed is the first acquired channel; policies start empty,
and this acquisition counts toward $k$.}
\label{fig:architecture}
\end{figure}

For ECG record $X_i$, an acquisition policy selects a lead set $S_i(k)$ with
$|S_i(k)|=k$ from the 12 standard channels. Evaluator $f$ maps the observed
waveforms and mask to five superclass probabilities. With diagnostic loss
$L_f(X_i,S)$, the value of another channel is
\begin{equation}
u_f(\ell\mid X_i,S)=L_f(X_i,S)-L_f(X_i,S\cup\{\ell\}).
\end{equation}
Thus, channel rankings can change when $f$ changes. Only acquired-channel
features enter the policy state. Both policies start empty; aVR is their first
acquisition and counts toward $k$ (Fig.~\ref{fig:architecture}a).

\clearpage
\onecolumn

\clearpage
\twocolumn[{\begin{minipage}{\textwidth}\centering
\includegraphics[width=\textwidth]{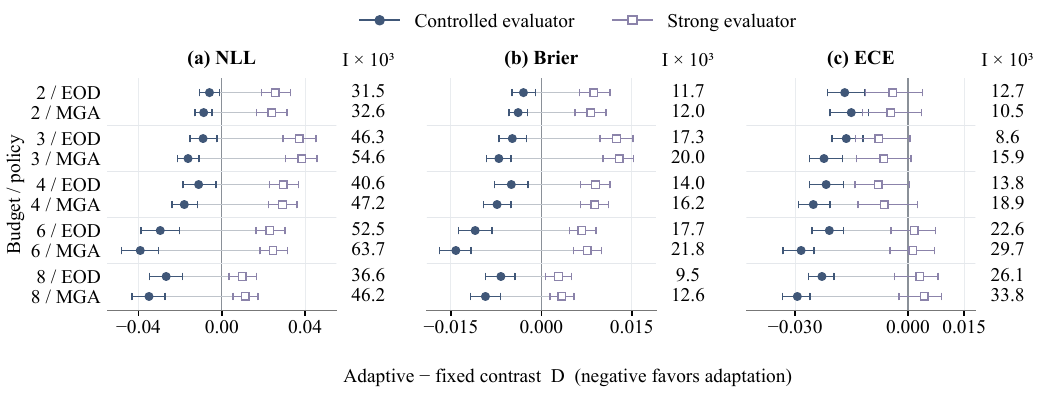}
\figcaption{Fold-8 adaptive-minus-fixed contrasts for NLL, Brier, and ECE.
Each row pairs the same policy and budget under controlled (filled circles)
and strong (open squares) evaluators, with evaluator-specific fixed references.
EOD denotes ECG-on-Demand. Whiskers show paired patient-cluster 95\% CIs;
gray segments connect estimates. Numeric columns report $10^3 I$.
Negative $D$ favors adaptation.}
\label{fig:transfer}
\vspace{10pt}
\end{minipage}}]
\subsection{Evaluators and policies}
The controlled evaluator combines ten statistics per available lead with a
mask-aware one-vs-rest logistic model trained on random exact-cardinality masks.
ECG-on-Demand uses acquired features, lead identities, the mask, current
probabilities, and remaining budget. Myopic Gain Acquisition (MGA) predicts
one-step controlled-evaluator negative log-likelihood (NLL) reduction.

The strong evaluator is a 2.77M-parameter masked ResNet1D trained on random
exact-cardinality masks. It receives ten-second waveforms with unavailable
standardized channels zeroed and a 12-bit mask at the prediction head.
It scores the original controlled-evaluator trajectories without regenerating them.

\subsection{Fixed references and uncertainty}
For evaluator $f$, metric $m$, and budget $k$, we evaluate all $\binom{12}{k}$
fixed subsets on the selection split and freeze the minimizer $F_{f,m,k}$.
Let $L_{f,m}(A_k)$ denote the metric over the policy's record-specific sets.
The adaptive-minus-fixed contrast and evaluator interaction are
\begin{align}
D_{f,m}(k)&=L_{f,m}(A_k)-L_{f,m}(F_{f,m,k}),\\
I_m(k)&=D_{\mathrm S,m}(k)-D_{\mathrm C,m}(k).
\label{eq:interaction}
\end{align}
Negative $D$ favors adaptation; positive $I$ indicates a less favorable
contrast after replacement. Common-reference analyses use the same frozen
$F_{\mathrm C}$ or $F_{\mathrm S}$ for both evaluators.

We report mean multilabel NLL, Brier score, and macro label-wise ten-bin
expected calibration error (ECE)~\cite{naeini2015bbq,guo2017calibration,vaicenavicius2019evaluating}.
AUROC and AUPRC describe discrimination. Confidence intervals (CIs) use 1,000
paired patient-cluster bootstrap replicates, with identical patient resamples
across policies, evaluators, and references.

\section{Experiments}
\subsection{Data splits and provenance}
PTB-XL v1.0.3 has 21,799 recordings from 18,869 patients; we use 100-Hz
signals and five superclasses~\cite{wagner2020ptbxl}. Folds 1--6 train models,
fold 7 supports validation, and fold 9 supports development audits.
Fold 10 selects fixed subsets and the original strong checkpoint by random-mask
NLL. Fold 8 (2,173 records, 1,881 patients) supplies reported evaluation.
Historical fold-8 exposure makes this analysis exploratory, despite retraining
reported models without it. Budgets are
$k\in\{2,3,4,6,8\}$.

\subsection{Frozen-policy evaluator replacement}
Each policy beats its controlled fixed comparator in all 15 fold-9 comparisons.
The strong evaluator improves full-lead fold-9 NLL from 0.519 to 0.276,
yet does not preserve the adaptive advantage.
On fold 8 at $k=4$, ECG-on-Demand changes from $D_{\mathrm C,NLL}=-0.0111$
to $D_{\mathrm S,NLL}=+0.0295$, giving $I=+0.0406$
(95\% CI $[+0.0303,+0.0499]$; paired contrasts in Fig.~\ref{fig:transfer}).

All 20 NLL and Brier mean contrasts reverse sign. ECE remains negative at
$k\in\{2,3,4\}$ and turns positive at larger budgets. All 30 interaction
CIs exclude zero in the positive direction. These correlated, pointwise
intervals do not constitute a simultaneous confidence statement.

\clearpage
\twocolumn[{\begin{minipage}{\textwidth}\centering
\includegraphics[width=\textwidth]{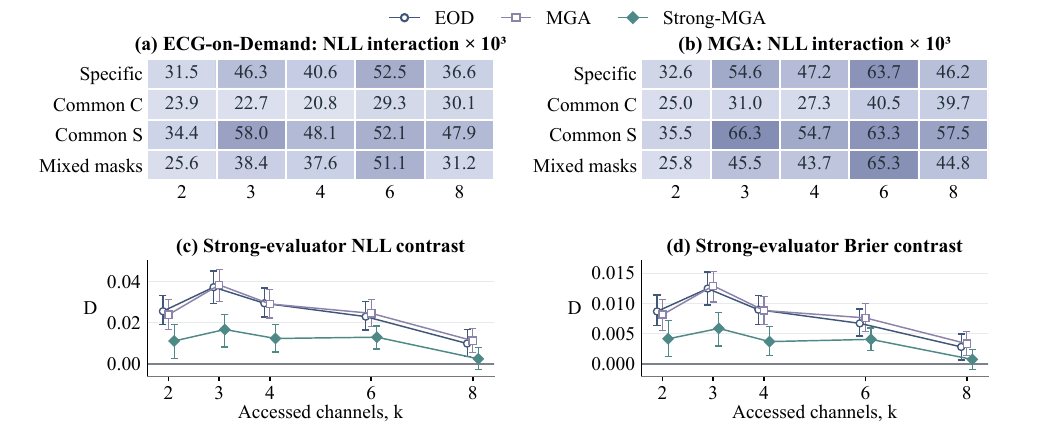}
\figcaption{Post-hoc fold-8 robustness and alignment. (a,b) NLL interactions
across budgets (columns) and reference/mask conditions (rows); cells show
$10^3 I$ on a shared color scale. All 40 associated 95\% CIs exclude zero.
(c,d) Strong-evaluator NLL and Brier contrasts for three policies. Whiskers
show paired patient-cluster 95\% CIs for gaps to the fixed baseline, not
between-policy differences.}
\label{fig:robustness}
\vspace{10pt}
\end{minipage}}]
\subsection{Reference and training-mask sensitivity}
These post-hoc analyses use fold 8 only for final scoring.
Both policies retain positive NLL interactions across budgets under either
common reference (Fig.~\ref{fig:robustness}(a,b)). At $k=4$, ECG-on-Demand yields
$+0.0208$ with $F_{\mathrm C}$ and $+0.0481$ with $F_{\mathrm S}$;
both 95\% CIs exclude zero. The reference changes magnitude, not direction.

We also train a strong evaluator on random masks (50\%),
frozen ECG-on-Demand masks (25\%), and frozen MGA masks (25\%). Training uses
folds 1--6 and budgets $\{1,2,3,4,6,8,12\}$; fold 7 selects the epoch, and
fold 10 provides a boundary check. Fold 8 is excluded from model development.
At $k=4$, the ECG-on-Demand NLL interaction remains
$+0.0376$ (95\% CI $[+0.0269,+0.0471]$). All 30 mixed-mask interaction
intervals exclude zero: broader mask exposure does not remove the shift.

\subsection{Evaluator-aligned policy training}
Strong-MGA learns one-step NLL reductions from the frozen original strong
evaluator, using training-record states and the same multilayer-perceptron
form and hyperparameters as controlled-trained MGA. This tests policy
alignment, rather than joint end-to-end optimization.

At $k=4$, Strong-MGA reduces strong-evaluator NLL from 0.2952 to 0.2784
relative to controlled-trained MGA (paired difference $-0.0168$, 95\% CI
$[-0.0260,-0.0084]$). Its adaptive-minus-fixed gap nevertheless remains
positive at $+0.0124$ ($[+0.0058,+0.0192]$; Fig.~\ref{fig:robustness}c).
This closes 57.6\% of MGA's original NLL gap at $k=4$, based on point estimates.
The Brier gap also narrows, while ECE shows no clear pairwise change.
Across budgets (Fig.~\ref{fig:robustness}c), aligned MGA has smaller mean NLL gaps;
the residual interval excludes zero at $k\in\{2,3,4,6\}$ but spans zero at
$k=8$. The latter does not establish equivalence to the fixed baseline.

\section{Discussion}
Better absolute prediction need not preserve an adaptive advantage: a stronger
evaluator may benefit its best fixed subset more than a transferred policy.
Channel-alignment analysis (Supplementary S8) supports this interpretation.
Adaptive sets favor controlled-evaluator leads and differ from the strong
evaluator's NLL-optimal set
$\{$III, aVR, V2, V5$\}$ (mean fold-8 Jaccard similarity 0.300).
Reference and mask checks weaken alternative explanations without identifying
a unique mechanism.

Myopic targets, limited policy capacity, or trajectory shift may constrain
Strong-MGA's recovery. Calibration requires separate assessment from NLL and
Brier. These retrospective, model- and dataset-specific results establish
neither clinical utility nor reduced electrode-placement burden. Historical
fold-8 exposure and post-hoc analyses limit confirmatory interpretation.

\section{Conclusion}
Replacing the diagnostic evaluator can reverse adaptive ECG advantages.
Reference and mask checks support this dependence, while aligned training
provides partial recovery. Acquisition policies should be validated with
their intended downstream model; joint optimization remains open.

\clearpage
\fontsize{9.2}{11.5}\selectfont
\section*{\normalsize COMPLIANCE WITH ETHICAL STANDARDS}
This study uses only public, de-identified PTB-XL data under its released terms; no intervention, enrollment, or new protected health information was collected.
\bibliographystyle{IEEEbib}
\bibliography{refs}

\begin{thebibliography}{10}

\bibitem{kligfield2007recommendations}
P.~Kligfield et~al.,
\newblock ``Recommendations for the standardization and interpretation of the
  electrocardiogram: Part {I}: The electrocardiogram and its technology,''
\newblock {\em Journal of the American College of Cardiology}, vol. 49, no. 10,
  pp. 1109--1127, 2007.

\bibitem{nelwan2004reconstruction}
S.~P. Nelwan, J.~A. Kors, S.~H. Meij, J.~H. van Bemmel, and M.~L. Simoons,
\newblock ``Reconstruction of the 12-lead electrocardiogram from reduced lead
  sets,''
\newblock {\em Journal of Electrocardiology}, vol. 37, no. 1, pp. 11--18, 2004.

\bibitem{reyna2021will}
M.~A. Reyna et~al.,
\newblock ``Will two do? varying dimensions in electrocardiography: The
  {PhysioNet}/computing in cardiology challenge 2021,''
\newblock in {\em Computing in Cardiology}, 2021, vol.~48, pp. 1--4.

\bibitem{lai2021optimal}
C.~Lai, S.~Zhou, and N.~A. Trayanova,
\newblock ``Optimal {ECG}-lead selection increases generalizability of deep
  learning on {ECG} abnormality classification,''
\newblock {\em Philosophical Transactions of the Royal Society A}, vol. 379,
  no. 2212, pp. 20200258, 2021.

\bibitem{oh2022leadagnostic}
J.~Oh, H.~Chung, J.~m.~Kwon, D.~g.~Hong, and E.~Choi,
\newblock ``Lead-agnostic self-supervised learning for local and global
  representations of electrocardiogram,''
\newblock in {\em Proceedings of the Conference on Health, Inference, and
  Learning}, 2022, vol. 174 of {\em Proceedings of Machine Learning Research},
  pp. 338--353.

\bibitem{chen2024mcma}
J.~Chen, W.~Wu, T.~Liu, and S.~Hong,
\newblock ``Multi-channel masked autoencoder and comprehensive evaluations for
  reconstructing 12-lead {ECG} from arbitrary single-lead {ECG},''
\newblock {\em npj Cardiovascular Health}, vol. 1, pp. 34, 2024.

\bibitem{iwata2024switching}
T.~Iwata, R.~Nishikimi, R.~Shibue, M.~Nakano, K.~Kashino, and H.~Tomoike,
\newblock ``Electrocardiographic classification using deep learning with lead
  switching,''
\newblock in {\em 2024 46th Annual International Conference of the IEEE
  Engineering in Medicine and Biology Society}, 2024, pp. 1--4.

\bibitem{ribeiro2020automatic}
A.~H. Ribeiro et~al.,
\newblock ``Automatic diagnosis of the 12-lead {ECG} using a deep neural
  network,''
\newblock {\em Nature Communications}, vol. 11, pp. 1760, 2020.

\bibitem{strodthoff2021deep}
N.~Strodthoff, P.~Wagner, T.~Schaeffter, and W.~Samek,
\newblock ``Deep learning for {ECG} analysis: Benchmarks and insights from
  {PTB-XL},''
\newblock {\em IEEE Journal of Biomedical and Health Informatics}, vol. 25, no.
  5, pp. 1519--1528, 2021.

\bibitem{shim2018pay}
H.~Shim, S.~J. Hwang, and E.~Yang,
\newblock ``Why pay more when you can pay less: A joint learning framework for
  active feature acquisition and classification,''
\newblock {\em arXiv preprint arXiv:1709.05964}, 2017.

\bibitem{janisch2019costly}
J.~Janisch, T.~Pevn{\'y}, and V.~Lis{\'y},
\newblock ``Classification with costly features using deep reinforcement
  learning,''
\newblock in {\em Proceedings of the AAAI Conference on Artificial
  Intelligence}, 2019, vol.~33, pp. 3959--3966.

\bibitem{ma2019eddi}
C.~Ma, S.~Tschiatschek, K.~Palla, J.~M. Hernandez-Lobato, S.~Nowozin, and
  C.~Zhang,
\newblock ``{EDDI}: Efficient dynamic discovery of high-value information with
  partial {VAE},''
\newblock in {\em Proceedings of the 36th International Conference on Machine
  Learning}, 2019, vol.~97 of {\em Proceedings of Machine Learning Research},
  pp. 4234--4243.

\bibitem{li2020acflow}
Y.~Li, S.~Akbar, and J.~Oliva,
\newblock ``{ACFlow}: Flow models for arbitrary conditional likelihoods,''
\newblock in {\em Proceedings of the 37th International Conference on Machine
  Learning}, 2020, vol. 119 of {\em Proceedings of Machine Learning Research},
  pp. 5831--5841.

\bibitem{gong2019icebreaker}
W.~Gong, S.~Tschiatschek, S.~Nowozin, R.~E. Turner, J.~M. Hernandez-Lobato, and
  C.~Zhang,
\newblock ``Icebreaker: Element-wise efficient information acquisition with a
  bayesian deep latent gaussian model,''
\newblock in {\em Advances in Neural Information Processing Systems}, 2019,
  vol.~32.

\bibitem{covert2023dynamic}
I.~C. Covert, W.~Qiu, M.~Lu, N.~Y. Kim, N.~J. White, and S.-I. Lee,
\newblock ``Learning to maximize mutual information for dynamic feature
  selection,''
\newblock in {\em Proceedings of the 40th International Conference on Machine
  Learning}, 2023, vol. 202 of {\em Proceedings of Machine Learning Research},
  pp. 6424--6447.

\bibitem{duarte2013taskdriven}
J.~M. Duarte-Carvajalino, G.~Yu, L.~Carin, and G.~Sapiro,
\newblock ``Task-driven adaptive statistical compressive sensing of gaussian
  mixture models,''
\newblock {\em IEEE Transactions on Signal Processing}, vol. 61, no. 3, pp.
  585--600, 2013.

\bibitem{chakrabarti2016sensor}
A.~Chakrabarti,
\newblock ``Learning sensor multiplexing design through back-propagation,''
\newblock in {\em Advances in Neural Information Processing Systems}, 2016,
  vol.~29.

\bibitem{vancalster2019calibration}
B.~Van Calster, D.~J. McLernon, M.~van Smeden, L.~Wynants, and E.~W.
  Steyerberg,
\newblock ``Calibration: The achilles heel of predictive analytics,''
\newblock {\em BMC Medicine}, vol. 17, pp. 230, 2019.

\bibitem{naeini2015bbq}
M.~P. Naeini, G.~Cooper, and M.~Hauskrecht,
\newblock ``Obtaining well calibrated probabilities using bayesian binning,''
\newblock in {\em Proceedings of the AAAI Conference on Artificial
  Intelligence}, 2015, vol.~29.

\bibitem{guo2017calibration}
C.~Guo, G.~Pleiss, Y.~Sun, and K.~Q. Weinberger,
\newblock ``On calibration of modern neural networks,''
\newblock in {\em Proceedings of the 34th International Conference on Machine
  Learning}, 2017, vol.~70, pp. 1321--1330.

\bibitem{vaicenavicius2019evaluating}
J.~Vaicenavicius, D.~Widmann, C.~Andersson, F.~Lindsten, J.~Roll, and
  T.~Sch{\"o}n,
\newblock ``Evaluating model calibration in classification,''
\newblock in {\em Proceedings of the Twenty-Second International Conference on
  Artificial Intelligence and Statistics}, 2019, vol.~89 of {\em Proceedings of
  Machine Learning Research}, pp. 3459--3467.

\bibitem{wagner2020ptbxl}
P.~Wagner, N.~Strodthoff, R.-D. Bousseljot, D.~Kreiseler, F.~I. Lunze,
  W.~Samek, and T.~Schaeffter,
\newblock ``{PTB-XL}, a large publicly available electrocardiography dataset,''
\newblock {\em Scientific Data}, vol. 7, no. 1, pp. 154, 2020.

\end{thebibliography}
\end{document}


\pdfinfo{/Title (Evaluator-Dependent Patient-Adaptive ECG Lead-Channel Allocation - Supplementary Material) /Author (Xiaoyang Li; Zeyan Tao)}
\pagestyle{plain}

\begin{center}
{\LARGE Evaluator-Dependent Patient-Adaptive ECG Lead-Channel Allocation\par}
\vspace{5pt}
{\large Supplementary Material\par}
\vspace{7pt}
Xiaoyang Li and Zeyan Tao\\
College of Medicine and Biological Information Engineering\\
Northeastern University, Shenyang, China\\
Xiaoyang Li: 20246389@stu.neu.edu.cn, ORCID: 0009-0002-4863-5761\\
Zeyan Tao: taozy@mails.neu.edu.cn, ORCID: 0009-0000-3149-9105
\end{center}
\vspace{5pt}

These supplementary analyses detail the experimental protocol and the
evidence underlying the main paper.  They cover the two diagnostic evaluators,
the acquisition comparators, exhaustive metric-matched fixed-subset search,
calibration diagnostics, patient-level heterogeneity, the evaluator
interaction, and post-hoc robustness analyses.  Results from the
fold-9 development set are identified as development-only and distinguished
from the held-out fold-8 evaluation.
Throughout, $k$ denotes the number of accessed ECG lead channels, not the
number of electrodes or independent electrode-placement actions.

\section*{S1. Cohort definition and analysis partitions}

The study uses PTB-XL v1.0.3, comprising 21,799 ten-second, 12-channel
recordings sampled at 100~Hz from 18,869 patients.  Outcomes are the five
diagnostic superclasses: conduction disturbance (CD), hypertrophy (HYP),
myocardial infarction (MI), normal ECG (NORM), and ST/T change (STTC).
Table~\ref{tab:splits} gives the official-fold roles used for the current analysis.
No patient occurs in more than one partition.

\begin{table}[H]
\centering
\caption{Frozen patient-disjoint PTB-XL analysis roles.  Records with no mapped
positive superclass remain in all analyses and are listed in the last column.}
\label{tab:splits}
\small
\begin{tabular}{lcccc}
\toprule
Role & Official fold(s) & Records & Patients & All-zero targets \\
\midrule
Training & 1--6 & 13,069 & 11,271 & 248 \\
Validation & 7 & 2,176 & 1,871 & 42 \\
Held-out evaluation & 8 & 2,173 & 1,881 & 44 \\
Development audit & 9 & 2,183 & 1,942 & 37 \\
Selection & 10 & 2,198 & 1,904 & 40 \\
\midrule
Total & 1--10 & 21,799 & 18,869 & 411 \\
\bottomrule
\end{tabular}
\end{table}

Before the boundary reset, fold 8 had appeared in historical training pools.
We therefore use the qualified term \emph{held-out exploratory evaluation}.
Models in the reported analysis were retrained without fold 8.  Record-level prevalence on
the held-out fold-8 set was 22.64\% for CD, 12.33\% for HYP, 24.76\% for MI,
42.89\% for NORM, and 23.65\% for STTC.

\section*{S2. Controlled evaluator and outcome definitions}

The controlled evaluator separates channel allocation from high-capacity
representation learning.  Ten deterministic waveform statistics describe
each available channel, and a mask-aware, balanced one-vs-rest logistic model
predicts the five superclasses.  Random exact-cardinality masking during
training supports arbitrary observed-channel subsets.  Adaptive and fixed
methods share the same fitted evaluator, with no method-specific or
post-hoc calibration.

For records $i=1,\ldots,N$ and labels $c=1,\ldots,5$, the frozen proper scores
are
\begin{align}
\mathrm{NLL}
&=-\frac{1}{5N}\sum_{i,c}
\left[y_{ic}\log p_{ic}+(1-y_{ic})\log(1-p_{ic})\right],\\
\mathrm{Brier}
&=\frac{1}{5N}\sum_{i,c}(p_{ic}-y_{ic})^2.
\end{align}
Macro ECE averages five label-wise ECE values computed from ten equal-width
probability bins.  Results with five, 15, and 20 bins are sensitivity analyses
only.  AUROC and AUPRC serve as descriptive discrimination measures rather
than substitutes for the probability-loss endpoints.  Point estimates are
globally record-weighted unless patient-equal aggregation is stated explicitly.

\section*{S3. Acquisition policies and comparators}

ECG-on-Demand scores the unobserved channels from the partial waveform state,
the acquisition mask, current controlled-evaluator probabilities, and the
remaining budget.  It is trained with a listwise action-ranking objective and
acquires exactly $k$ channels at inference, without a stop action.  The frozen
ECG-on-Demand selector uses $\lambda=0$ for the redundancy penalty.

Myopic Gain Acquisition (MGA) provides a learned, budget-agnostic adaptive comparator.
From the current partial state, it estimates the one-step marginal BCE
reduction for each unavailable channel and greedily selects the largest
predicted gain until exactly $k$ channels have been observed.
Cohort-Conditional Greedy Acquisition (CCGA) supplies a non-neural adaptive comparator,
mapping the current diagnostic state to conditional marginal gains estimated
from the training cohort.  At inference, none of these methods has access to
labels, unobserved waveforms, or oracle utilities.  Since no patient waveform
is available at $t=0$, all policies begin with the same population-wide action,
aVR.

Two further development controls distinguish state conditioning from subset
diversity.  The cohort-state control substitutes the cohort-average state for
the patient-specific state; the random-subset control draws patient-specific
exact-cardinality subsets without consulting the observed state.  Neither
control contributes to fixed-subset selection or the primary interaction.

At $k=4$, the fold-9 ECG-on-Demand-minus-MGA intervals included zero for NLL,
Brier, and ECE, whereas CCGA yielded higher values on all three metrics
(Table~\ref{tab:adaptive-comparators}).  The two learned methods nevertheless
followed different trajectories: only 4.67\% of their $k=4$ final subsets were
identical, with mean final-set Jaccard similarity 0.427.  The result favors
learned state-dependent allocation but does not establish architectural
superiority.

\begin{table}[H]
\centering
\caption{Development-only $k=4$ adaptive-policy comparisons under the
controlled evaluator.  Each delta is ECG-on-Demand minus the named comparator;
negative values favor ECG-on-Demand.}
\label{tab:adaptive-comparators}
\small
\begin{tabular}{llrrr}
\toprule
Comparator & Metric & ECG-on-Demand & Comparator & Delta [95\% CI] \\
\midrule
MGA & NLL & .544111 & .541128 & +.002983 [$-.002325,+.008198$] \\
MGA & Brier & .177615 & .177582 & +.000032 [$-.002090,+.002137$] \\
MGA & ECE & .171480 & .171504 & $-.000025$ [$-.004108,+.002588$] \\
\addlinespace
CCGA & NLL & .544111 & .577556 & $-.033445$ [$-.042057,-.024645$] \\
CCGA & Brier & .177615 & .191016 & $-.013401$ [$-.015897,-.010965$] \\
CCGA & ECE & .171480 & .195674 & $-.024194$ [$-.028501,-.020903$] \\
\bottomrule
\end{tabular}
\end{table}

At $k=6$ and $k=8$, MGA achieved lower NLL, Brier, and ECE than
ECG-on-Demand, and all six paired intervals excluded zero.  The comparison
therefore provides no evidence that ListNet ranking or explicit budget
conditioning is necessary or dominant.

\section*{S4. Metric-matched fixed-subset search}

Population-wide fixed comparators were selected exclusively on fold 10.  For
each evaluator and budget in $\{1,2,3,4,6,8\}$, the search scored every
$\binom{12}{k}$ subset and separately selected the NLL-, Brier-, and
ECE-optimal configurations.  Deterministic tie-breaking used the target
metric, followed by NLL and lexicographic channel-index order.  Neither fold 9
nor fold 8 contributed to selection.

\begin{table}[H]
\centering
\caption{Exhaustive search size for each exact budget.}
\label{tab:search-size}
\small
\begin{tabular}{lrrrrrrr}
\toprule
$k$ & 1 & 2 & 3 & 4 & 6 & 8 & Total \\
\midrule
Candidates & 12 & 66 & 220 & 495 & 924 & 495 & 2,212 \\
\bottomrule
\end{tabular}
\end{table}

\begin{table}[H]
\centering
\caption{Frozen metric-matched fixed subsets.  C and S denote the controlled
and strong evaluators.}
\label{tab:fixed-subsets}
\small
\begin{tabular}{rlll}
\toprule
$k$ & Metric & Controlled fixed set & Strong fixed set \\
\midrule
1 & NLL & aVR & V6 \\
1 & Brier & aVR & V6 \\
1 & ECE & aVR & V4 \\
2 & NLL & aVR, aVL & III, aVR \\
2 & Brier & aVR, aVL & III, aVR \\
2 & ECE & aVR, V1 & aVL, V2 \\
3 & NLL & aVR, aVL, V4 & III, aVR, V2 \\
3 & Brier & aVR, aVL, V4 & III, aVR, V2 \\
3 & ECE & I, aVR, aVL & aVL, V2, V3 \\
4 & NLL & aVR, aVL, V3, V5 & III, aVR, V2, V5 \\
4 & Brier & aVR, aVL, V3, V5 & III, aVR, V2, V5 \\
4 & ECE & I, aVR, aVL, V5 & aVL, V1, V2, V3 \\
6 & NLL & I, aVR, aVL, V2, V4, V5 & aVR, aVL, aVF, V1, V2, V5 \\
6 & Brier & I, aVR, aVL, V2, V4, V5 & aVR, aVL, aVF, V1, V2, V5 \\
6 & ECE & I, II, aVR, aVL, V1, V5 & III, aVR, V1, V2, V3, V4 \\
8 & NLL & I, II, III, aVR, aVL, V2, V4, V5 & aVR, aVL, aVF, V1, V2, V3, V4, V5 \\
8 & Brier & I, II, III, aVR, aVL, V1, V3, V5 & aVR, aVL, aVF, V1, V2, V3, V4, V5 \\
8 & ECE & I, II, aVR, aVL, V1, V2, V4, V5 & I, aVR, aVL, V1, V2, V3, V4, V5 \\
\bottomrule
\end{tabular}
\end{table}

The controlled $k=1$ NLL-optimal channel, aVR, also serves as the
method-independent initial-uncertainty anchor for the patient analysis.

\section*{S5. Controlled-evaluator results on the fold-9 development set}

Table~\ref{tab:development-controlled} summarizes the development-only fold-9
analysis.  ECG-on-Demand yielded a negative adaptive-minus-fixed difference at
every budget and for every metric, with each paired interval below zero.  The
15 development contrasts provided context for the later held-out evaluation;
they are not holdout estimates themselves.

\begin{table}[H]
\centering
\caption{Development-only controlled-evaluator deltas against separately
metric-matched fixed subsets.}
\label{tab:development-controlled}
\small
\begin{tabular}{clll}
\toprule
$k$ & $\Delta$NLL [95\% CI] & $\Delta$Brier [95\% CI] & $\Delta$ECE [95\% CI] \\
\midrule
2 & $-.006814$ [$-.011139,-.002610$] & $-.003317$ [$-.005208,-.001571$] & $-.011113$ [$-.015931,-.007354$] \\
3 & $-.012196$ [$-.020452,-.005045$] & $-.005077$ [$-.007585,-.002759$] & $-.018295$ [$-.022982,-.013824$] \\
4 & $-.016554$ [$-.024120,-.007306$] & $-.008665$ [$-.011319,-.005845$] & $-.023655$ [$-.028406,-.019574$] \\
6 & $-.023518$ [$-.031969,-.015278$] & $-.008618$ [$-.011189,-.006079$] & $-.017815$ [$-.022445,-.015426$] \\
8 & $-.027341$ [$-.034889,-.020043$] & $-.007428$ [$-.009473,-.005076$] & $-.017846$ [$-.020853,-.015768$] \\
\bottomrule
\end{tabular}
\end{table}

At $k=4$, the absolute adaptive/fixed values were .544111/.560665 for NLL,
.177615/.186280 for Brier, and .171480/.195135 for ECE.  NLL and Brier are
compared with their respective proper-score-optimal fixed subsets rather than
with the ECE-optimal subset.

\section*{S6. Calibration and probability dispersion on the fold-9 development set}

\begin{figure}[H]
\centering
\includegraphics[width=\textwidth]{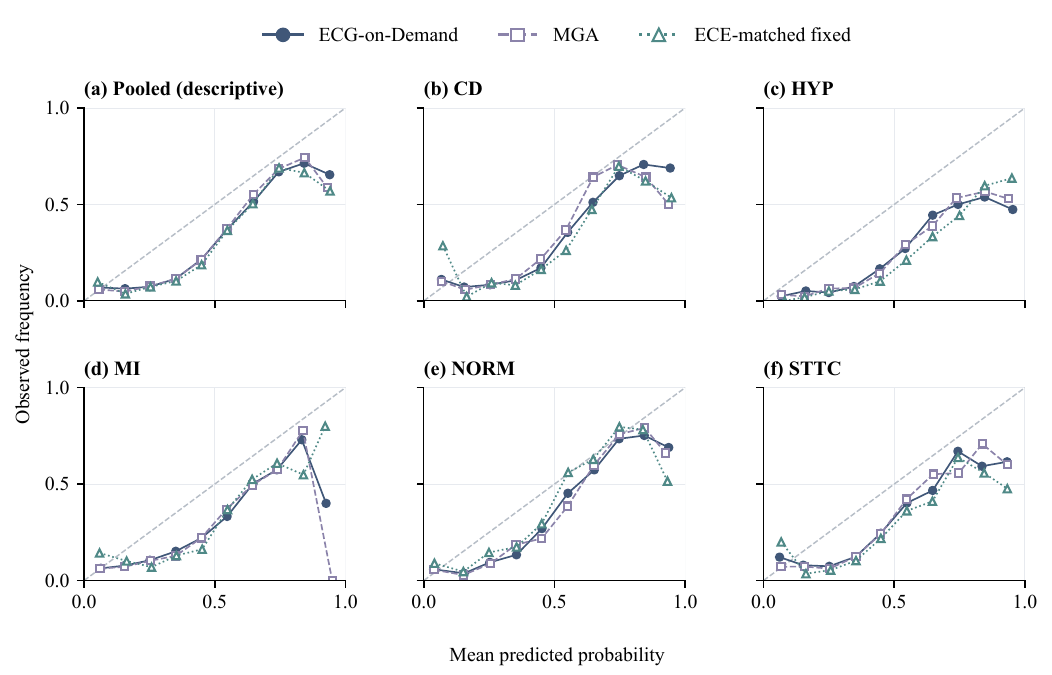}
\caption{Development-only reliability diagrams at $k=4$.  All methods share
the controlled evaluator and ten equal-width bins; only nonempty bins are
drawn.  Pooled is descriptive, and CD, HYP, MI, NORM, and STTC denote the five
PTB-XL diagnostic superclasses.  Label-wise inference is reported in the text,
not read from individual bin segments.}
\label{fig:calibration}
\end{figure}

Figure~\ref{fig:calibration} shows the development-only $k=4$ reliability
curves for ECG-on-Demand, MGA, and the ECE-matched fixed subset under common
bin definitions.  The pooled panel is descriptive, whereas the five
label-specific panels retain the prespecified macro label-wise ECE definition.

The ECE contrast remained negative with 5, 10, 15, and 20 bins: $-.021036$,
$-.023655$, $-.021181$, and $-.022803$, respectively.  Prediction variance
was .040482 for ECG-on-Demand and .037187 for the ECE-matched fixed subset,
corresponding to a retention ratio of 1.0886.  Mean predictive entropy was also
lower for ECG-on-Demand (.581565 versus .601231).  The dispersion measures
do not support collapse toward a narrow or
prevalence-like probability range as the explanation for lower ECE.

Calibration-in-the-large (CITL) is the fitted intercept with the logit
coefficient fixed at one; the calibration slope comes from a joint
intercept-and-slope fit.  Table~\ref{tab:calibration-structure} gives the macro
mean absolute CITL and absolute slope deviation from one.

\begin{table}[H]
\centering
\caption{Development-only $k=4$ calibration-structure summary.  The fixed
column is ECE-matched, so its NLL and Brier entries are descriptive rather than
the primary proper-score comparators.}
\label{tab:calibration-structure}
\small
\begin{tabular}{lrrr}
\toprule
Quantity & ECG-on-Demand & MGA & ECE-matched fixed \\
\midrule
NLL & .544111 & .541128 & .568881 \\
Brier & .177615 & .177582 & .187824 \\
ECE & .171480 & .171504 & .195135 \\
Macro mean $|\mathrm{CITL}|$ & .9858 & .9627 & 1.0716 \\
Macro mean $|\mathrm{slope}-1|$ & .0472 & .1807 & .1107 \\
Probability variance & .04048 & .03592 & .03719 \\
Predictive entropy & .5816 & .5930 & .6012 \\
\bottomrule
\end{tabular}
\end{table}

Label-wise ECG-on-Demand-minus-fixed ECE differences were $-.042872$ for CD,
$-.030019$ for HYP, $-.032742$ for MI, $+.027724$ for NORM, and $-.040368$
for STTC.  The NORM interval lay entirely above zero
([$+.005422,+.042757$]), whereas the remaining four intervals lay entirely
below zero.  Intercept and slope fits are diagnostic summaries, not applied
recalibration models; their absolute values should also be interpreted in
light of class-balanced logistic training.  The evidence supports lower
probability loss and lower macro label-wise ECE, not uniform improvement in
calibration structure.

\section*{S7. Patient-level heterogeneity on the fold-9 development set}

The development-only patient analysis uses the controlled evaluator at $k=4$
and the NLL-matched fixed subset.  Multilabel mean NLL is first computed for
each record and then averaged within patient, giving equal weight to each of
the 1,942 fold-9 patients.  We retain
$\Delta\mathrm{NLL}=\mathrm{adaptive}-\mathrm{fixed}$, so negative values
favor adaptive allocation.

\begin{figure}[H]
\centering
\includegraphics[width=\textwidth]{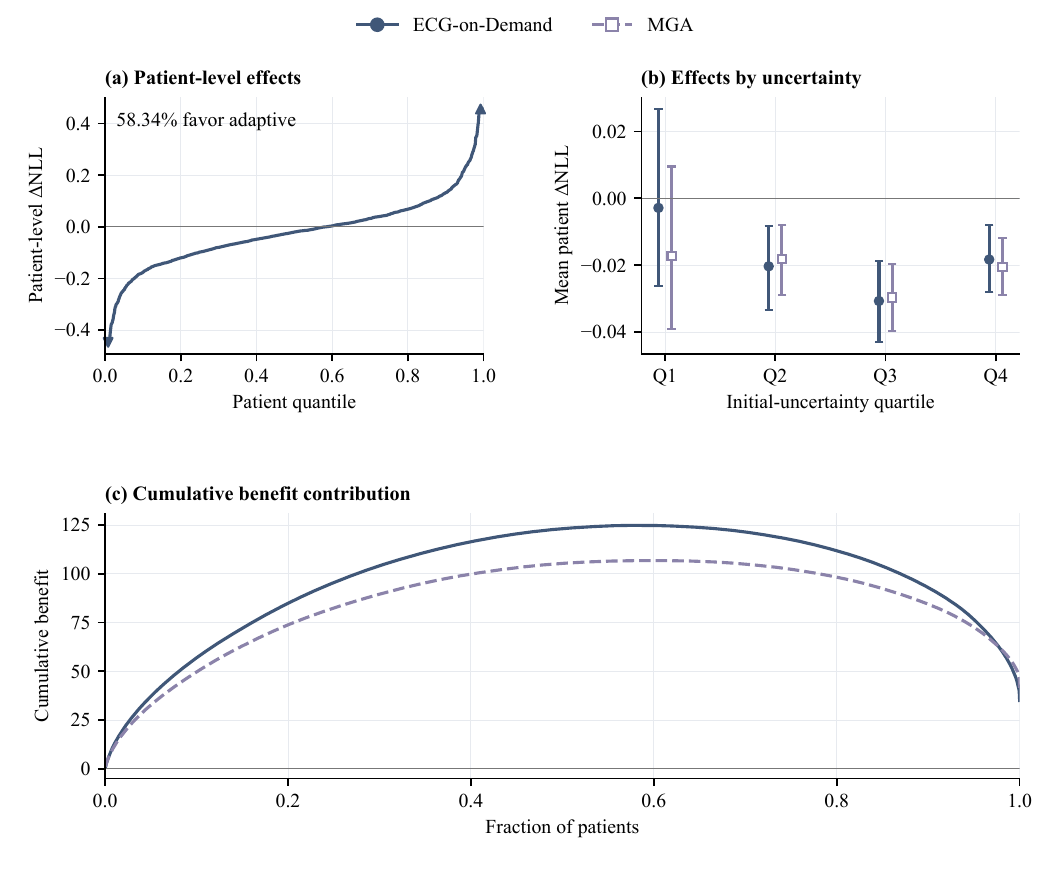}
\caption{Development-only patient heterogeneity.  (A) Sorted ECG-on-Demand
patient-level $\Delta$NLL; the axis displays the central 98\%, with the full
range extending from $-1.15$ to $+4.89$.  (B) Mean effects across uncertainty
quartiles after the common aVR anchor.  (C) Cumulative benefit after ranking
patients separately within each policy; horizontal positions do not identify
the same patient across policies.}
\label{fig:patient-heterogeneity}
\end{figure}

\begin{table}[H]
\centering
\caption{Prespecified patient-equal robustness summaries for ECG-on-Demand.}
\label{tab:patient-effects}
\small
\begin{tabular}{lrr}
\toprule
Statistic & Estimate & 95\% CI \\
\midrule
Mean $\Delta$NLL & $-.018060$ & [$-.026037,-.008932$] \\
Median $\Delta$NLL & $-.020121$ & [$-.026731,-.015466$] \\
5\% trimmed mean & $-.022540$ & [$-.028158,-.016543$] \\
10\% trimmed mean & $-.023193$ & [$-.028507,-.017958$] \\
5\% winsorized mean & $-.021231$ & [$-.027310,-.014704$] \\
Exclude largest 1\% $|\Delta|$ & $-.020970$ & [$-.027137,-.014538$] \\
Fraction improved & 58.34\% & [56.18\%,60.61\%] \\
Fraction worsened & 41.61\% & [39.34\%,43.82\%] \\
Fraction unchanged & 0.05\% & [0.00\%,0.15\%] \\
\bottomrule
\end{tabular}
\end{table}

The corresponding counts are 1,133 patients improved, 808 worsened, and one
was numerically unchanged.

The prespecified uncertainty association was not supported.  Spearman
$\rho(U,-\Delta\mathrm{NLL})$ was $-.0453$ ([$-.0916,-.0013$]) for
ECG-on-Demand and $-.0361$ ([$-.0867,+.0096$]) for MGA.  From Q1 to Q4,
ECG-on-Demand quartile means were $-.002878$, $-.020349$, $-.030723$, and
$-.018319$; Q4 also had the lowest benefit rate, 55.6\%.  The benefited-patient
sets had Jaccard similarity .612: 44.44\% of patients benefited under both
policies, 13.90\% only under ECG-on-Demand, 14.26\% only under MGA, and 27.39\%
under neither.

The label-aware subgroup descriptions are retrospective and exploratory.
Patients with one positive superclass had mean $\Delta$NLL $-.032594$,
whereas those with at least two had $+.025782$.  Mean values were $-.071302$
for NORM-positive patients and $+.028405$ for NORM-negative patients.  These
associations do not define a prospective patient-selection rule, a subgroup
policy, or a causal treatment effect.

\section*{S8. Strong raw-waveform evaluation}

The prespecified strong evaluator is a MaskedResNet1D with 2,774,685
parameters.  Its input is a 12-channel, 1,000-sample waveform with unavailable
channels zeroed, together with an explicit 12-dimensional acquisition mask.
Residual stages have widths 64, 96, 160, and 256, with two blocks per stage,
followed by temporal pooling and prediction of the same five superclasses.
Training on folds 1--6 used random exact-cardinality masks from
$k\in\{1,2,3,4,6,8,12\}$.  The same evaluator serves all policies and fixed
subsets, without policy-specific calibration.

\paragraph*{Mask distribution note.}
The ResNet1D was trained under random exact-cardinality masks drawn uniformly
from all $\binom{12}{k}$ subsets at each $k$.  At inference, adaptive policies
produce systematically non-random masks: ECG-on-Demand and MGA concentrate
selection on high-marginal-gain channels under the controlled evaluator (e.g.\
aVR, V3, V5 at $k=4$; see main-paper Fig.~1(c) and
Table~\ref{tab:fixed-subsets}).  Population-wide fixed subsets also produce
non-random masks, but they are homogeneous across patients, whereas adaptive
masks are patient-heterogeneous.  Whether the ResNet1D's response to
heterogeneous systematic masks differs from its response to a single fixed mask
is not directly testable without retraining; however, the strong-evaluator fixed
comparator was selected by scoring all $\binom{12}{k}$ subsets with the same
ResNet1D on fold 10, so any additive scoring artifact shared by adaptive and
fixed subsets under the same evaluator cancels in the contrast $\DS$.  The
interaction $\Ieval = \DS - \DC$ therefore captures the change in the
adaptive-versus-best-prespecified-fixed margin across evaluators rather than a
raw evaluator-level effect on the adaptive masks alone.

Selection among three prespecified seeds was based solely on
policy-independent random-mask NLL on fold 10.  Epoch 54 of seed 20060928 was
selected, with fold-7 checkpoint NLL .297597 and fold-10 selection NLL
.325522.  Acquisition trajectories were not regenerated: the strong evaluator
scored the channels chosen by the fixed controlled-evaluator policies.

\begin{table}[H]
\centering
\caption{Development-only full-12 evaluator quality on fold 9.}
\label{tab:evaluator-quality}
\small
\begin{tabular}{lrrrrr}
\toprule
Evaluator & NLL & Brier & ECE & AUROC & AUPRC \\
\midrule
Controlled statistics logistic & .519369 & .161953 & .157117 & .830021 & .598992 \\
Strong MaskedResNet1D & .275906 & .083566 & .028349 & .927667 & .808946 \\
\bottomrule
\end{tabular}
\end{table}

\paragraph*{Channel-set alignment with strong-evaluator optima.}
The strong-evaluator NLL-optimal fixed subset at $k=4$ is
$F_{S,\text{NLL},4}=\{\text{III},\text{aVR},\text{V2},\text{V5}\}$.
To probe the mechanism underlying the transfer failure, we compute the
Jaccard similarity between each held-out fold-8 ECG-on-Demand output set
$S_i(4)$ and $F_{S,\text{NLL},4}$ across all 2,173 fold-8 records.
The mean Jaccard similarity is 0.300 (range $[1/7, 3/5]$).  The
controlled-evaluator NLL-optimal fixed subset
$F_{C,\text{NLL},4}=\{\text{aVR},\text{aVL},\text{V3},\text{V5}\}$
shares aVR and V5 with the strong fixed set. The mean Jaccard similarity
to this controlled fixed set is 0.251; adaptive outputs therefore do not
concentrate on either fixed optimum.

Because aVR is the universal first-step anchor, it appears in 100\% of
$S_i(4)$.  The remaining three channels of $F_{S,\text{NLL},4}$---III,
V2, and V5---appear in only 28.1\%, 18.8\%, and 32.2\% of records,
respectively.  Leads I (55.1\%), II (49.2\%), V1 (37.4\%), and V6
(30.7\%) dominate the adaptive output instead; none of these belong to
$F_{S,\text{NLL},4}$.  Consequently, 29.8\% of records share only aVR
with $F_{S,\text{NLL},4}$ ($|S_i(4)\cap F_{S,\text{NLL},4}|=1$), 61.3\%
share two channels (aVR plus one of III, V2, or V5), and only 8.9\%
share three.  No record's $S_i(4)$ is identical to $F_{S,\text{NLL},4}$.
This pattern is consistent with evaluator-relative utility (Eq.~1 in
the main paper): leads I, II, V1, and V6 carry marginal gain under the
controlled statistics-logistic evaluator but are displaced by III and V2
in the raw-waveform convolutional feature space.  The low overlap between
adaptive outputs and $F_{S,\text{NLL},4}$ provides a channel-level
account of the strong-evaluator NLL reversal.  It does not establish
causality; a policy trained jointly with the strong evaluator might
converge to a different channel-preference profile.

Although more predictive in the full-12 development check, the raw-waveform model did not preserve the
controlled-trained allocation advantage on the fold-9 development set.  At $k=4$,
ECG-on-Demand minus the strong-evaluator-specific fixed comparator gave NLL
$+.032504$ [$+.025201,+.039939$], Brier $+.010348$
[$+.007947,+.012848$], and ECE $+.005011$ [$-.004008,+.009868$].  MGA gave
NLL $+.030503$, Brier $+.009609$, and ECE $+.002692$ under the same
metric-matching principle.  Across all five budgets, both learned policies had
positive strong-evaluator NLL and Brier differences with intervals above zero.
This experiment evaluates transfer of frozen policies.  The post-hoc
evaluator-aligned Strong-MGA test in S13 directly examines whether changing the
policy target can recover the strong-evaluator gap.

\section*{S9. Evaluator interaction on the held-out fold 8}

\begin{figure}[!b]
\centering
\includegraphics[width=\textwidth]{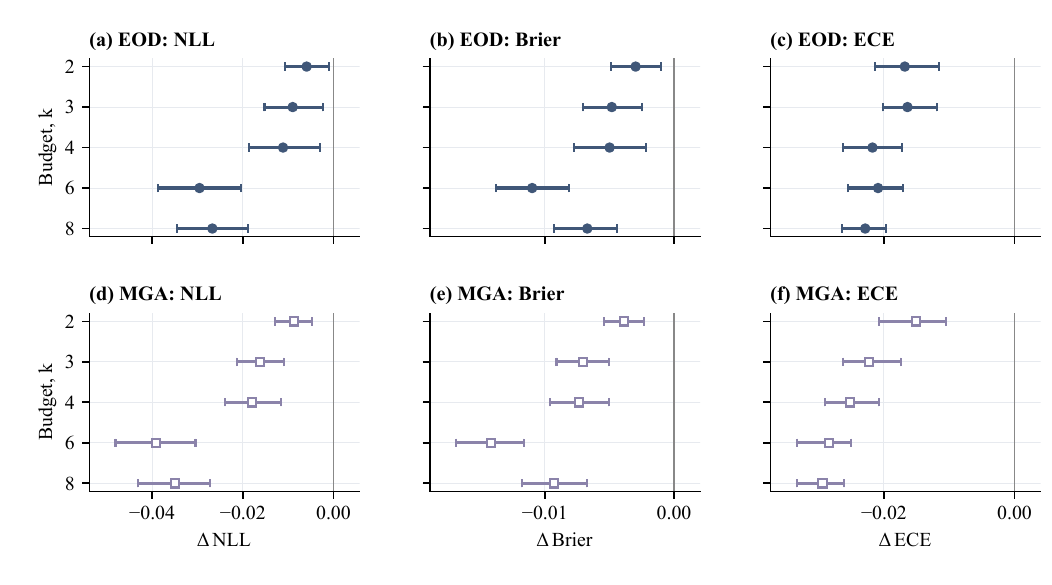}
\caption{Controlled-evaluator effects on the held-out fold-8 evaluation.
Points show adaptive minus separately selected metric-matched fixed values;
negative values favor the learned policy.  Error bars are policy-specific
paired 95\% patient-cluster bootstrap intervals computed from identical
patient resamples across ECG-on-Demand and MGA.}
\label{fig:controlled-locked}
\end{figure}

The held-out fold-8 evaluation comprises 2,173 records from 1,881
patients.  Under the controlled evaluator, both learned policies outperform
their metric-matched fixed comparators at the primary and secondary budgets
for every metric shown in Fig.~\ref{fig:controlled-locked}.

\begin{table}[H]
\centering
\caption{Primary $k=4$ held-out results.  Here $\DC$ and $\DS$ are
adaptive minus the evaluator-specific metric-matched fixed comparator, and
$\Ieval=\DS-\DC$.}
\label{tab:locked-k4}
\small
\begin{tabular}{lrrrr}
\toprule
Metric & C adaptive & C fixed & S adaptive & S fixed \\
\midrule
NLL & .537447 & .548567 & .295501 & .266035 \\
Brier & .176673 & .181657 & .090714 & .081737 \\
ECE & .174548 & .196308 & .016474 & .024393 \\
\bottomrule\end{tabular}
\par\vspace{8pt}
\begin{tabular}{lll}
\toprule
Metric & Contrast & Estimate [95\% CI] \\
\midrule
NLL & $\DC$ & $-.011120$ [$-.018652,-.002885$] \\
NLL & $\DS$ & $+.029466$ [$+.022848,+.036799$] \\
NLL & $\Ieval$ & $+.040585$ [$+.030288,+.049879$] \\
Brier & $\DC$ & $-.004984$ [$-.007763,-.002193$] \\
Brier & $\DS$ & $+.008977$ [$+.006516,+.011366$] \\
Brier & $\Ieval$ & $+.013960$ [$+.010498,+.017154$] \\
ECE & $\DC$ & $-.021761$ [$-.026244,-.017206$] \\
ECE & $\DS$ & $-.007919$ [$-.014115,+.000486$] \\
ECE & $\Ieval$ & $+.013841$ [$+.006892,+.023210$] \\
\bottomrule\end{tabular}
\end{table}

\begin{table}[H]
\centering
\caption{Complete held-out ECG-on-Demand evaluator interactions.}
\label{tab:locked-ecgod}
\small
\begin{tabular}{rllll}
\toprule
$k$ & Metric & $\DC$ & $\DS$ & $\Ieval$ [95\% CI] \\
\midrule
2 & NLL & $-.005916$ & $+.025592$ & $+.031508$ [$+.023812,+.039370$] \\
2 & Brier & $-.002970$ & $+.008690$ & $+.011660$ [$+.009020,+.014504$] \\
2 & ECE & $-.016806$ & $-.004078$ & $+.012728$ [$+.004011,+.021152$] \\
3 & NLL & $-.008996$ & $+.037281$ & $+.046277$ [$+.037224,+.055788$] \\
3 & Brier & $-.004814$ & $+.012458$ & $+.017272$ [$+.013839,+.020392$] \\
3 & ECE & $-.016399$ & $-.007769$ & $+.008630$ [$+.001107,+.017514$] \\
4 & NLL & $-.011120$ & $+.029466$ & $+.040585$ [$+.030288,+.049879$] \\
4 & Brier & $-.004984$ & $+.008977$ & $+.013960$ [$+.010498,+.017154$] \\
4 & ECE & $-.021761$ & $-.007919$ & $+.013841$ [$+.006892,+.023210$] \\
6 & NLL & $-.029544$ & $+.022986$ & $+.052530$ [$+.041111,+.063360$] \\
6 & Brier & $-.010976$ & $+.006722$ & $+.017698$ [$+.014249,+.021098$] \\
6 & ECE & $-.020905$ & $+.001688$ & $+.022593$ [$+.015764,+.029835$] \\
8 & NLL & $-.026713$ & $+.009912$ & $+.036625$ [$+.026408,+.046312$] \\
8 & Brier & $-.006709$ & $+.002840$ & $+.009549$ [$+.006534,+.012696$] \\
8 & ECE & $-.022873$ & $+.003206$ & $+.026079$ [$+.018554,+.032258$] \\
\bottomrule
\end{tabular}
\end{table}

\begin{table}[H]
\centering
\caption{Complete held-out MGA evaluator interactions.}
\label{tab:locked-mga}
\small
\begin{tabular}{rllll}
\toprule
$k$ & Metric & $\DC$ & $\DS$ & $\Ieval$ [95\% CI] \\
\midrule
2 & NLL & $-.008706$ & $+.023940$ & $+.032646$ [$+.025151,+.040349$] \\
2 & Brier & $-.003855$ & $+.008165$ & $+.012020$ [$+.009186,+.014698$] \\
2 & ECE & $-.015065$ & $-.004572$ & $+.010494$ [$+.001993,+.020145$] \\
3 & NLL & $-.016210$ & $+.038374$ & $+.054584$ [$+.045825,+.062915$] \\
3 & Brier & $-.007046$ & $+.012913$ & $+.019959$ [$+.016900,+.022854$] \\
3 & ECE & $-.022325$ & $-.006400$ & $+.015925$ [$+.007074,+.023868$] \\
4 & NLL & $-.018011$ & $+.029139$ & $+.047150$ [$+.038922,+.055669$] \\
4 & Brier & $-.007348$ & $+.008845$ & $+.016192$ [$+.013386,+.019007$] \\
4 & ECE & $-.025176$ & $-.006268$ & $+.018908$ [$+.011009,+.027775$] \\
6 & NLL & $-.039141$ & $+.024586$ & $+.063728$ [$+.053703,+.074156$] \\
6 & Brier & $-.014181$ & $+.007625$ & $+.021806$ [$+.018695,+.025142$] \\
6 & ECE & $-.028419$ & $+.001320$ & $+.029739$ [$+.023312,+.037169$] \\
8 & NLL & $-.034944$ & $+.011281$ & $+.046225$ [$+.037196,+.055454$] \\
8 & Brier & $-.009274$ & $+.003373$ & $+.012647$ [$+.009673,+.015801$] \\
8 & ECE & $-.029426$ & $+.004377$ & $+.033804$ [$+.026251,+.039493$] \\
\bottomrule
\end{tabular}
\end{table}

All 15 ECG-on-Demand interaction intervals and all 15 MGA intervals lie above
zero.  Within each policy, every controlled-evaluator difference is negative
with its interval below zero, whereas every strong-evaluator NLL and Brier
difference is positive with its interval above zero.  Strong-evaluator ECE is
negative at $k=2,3,4$ and positive at $k=6,8$, but each of the five
evaluator-specific ECE intervals crosses zero.  The interaction therefore
shows a consistent evaluator-associated shift in the adaptive-versus-fixed
contrast without implying an ECE reversal at every budget.

\section*{S10. Original interaction and statistical inference}

For the original primary $k=4$ NLL analysis, $A$ denotes the originally specified
ECG-on-Demand policy, $F_{\mathrm C}$ the controlled-evaluator NLL-optimal
fixed subset, and $F_{\mathrm S}$ the strong-evaluator NLL-optimal fixed
subset.  The evaluator-specific contrasts and their interaction are
\begin{align}
\DC &= \mathrm{NLL}_{\mathrm C}(A)-\mathrm{NLL}_{\mathrm C}(F_{\mathrm C}),\\
\DS &= \mathrm{NLL}_{\mathrm S}(A)-\mathrm{NLL}_{\mathrm S}(F_{\mathrm S}),\\
\Ieval &= \DS-\DC.
\end{align}
Negative $\DC$ or $\DS$ favors adaptive allocation within the corresponding
evaluator.  The original hypothesis, $\Ieval>0$, specified a less favorable
adaptive-versus-fixed contrast under the strong evaluator.  The original
support criterion required the paired 95\% interval for $\Ieval$ to remain above
zero.  The primary interaction was $+.040585$ [$+.030288,+.049879$].  NLL
also exhibited the ordered reversal: the controlled interval was below zero,
whereas the strong interval was above zero.

Because $F_{\mathrm C}$ and $F_{\mathrm S}$ were selected separately on fold
10 before holdout access, $\Ieval$ asks whether the frozen adaptive policy
retains an advantage over the best prespecified population-wide protocol for
the corresponding evaluator.  It does not isolate an evaluator-only effect on
the adaptive masks independently of the changing fixed reference.

The Brier interaction was the key supporting endpoint, with ten-bin macro
label-wise ECE designated as secondary.  MGA and $k=2,3,6,8$ were included as
secondary policy and budget analyses in the original study.  The design introduced no
equivalence or noninferiority margin, subgroup-selection rule, or clinical
threshold.

Intervals are based on exactly 1,000 paired patient-cluster bootstrap
replicates with seed 20270909 and the 2.5th/97.5th percentile convention.  All
records from a sampled patient retain the same multiplicity, and every policy
and evaluator-specific fixed comparator uses the same resample.  No
record-level bootstrap was used.
Evaluator parameters, trajectories, fixed subsets, preprocessing, endpoints,
budgets, metric definitions, and interval conventions for the original
frozen-policy analysis were fixed before fold-8 scoring.  The common-reference,
mask-robust, and Strong-MGA analyses were added post hoc; they use the same
roles and bootstrap convention but are not confirmatory extensions.

\section*{S11. Common-reference sensitivity}

The original interaction uses evaluator-specific fixed references, so a
reference change could in principle contribute to the observed shift.  We
therefore froze the original $F_{\mathrm C}$ and $F_{\mathrm S}$ subsets and
scored both evaluators against each common reference.  At $k=4$ for
ECG-on-Demand, the three NLL estimands are:

\begin{table}[H]
\centering
\caption{Common-reference sensitivity for ECG-on-Demand at $k=4$ on the
held-out fold-8 evaluation.  Negative $D$ favors adaptive allocation.}
\small
\begin{tabular}{lrrr}
\toprule
Reference & $D_{\mathrm C}$ & $D_{\mathrm S}$ & $I^{\mathrm{ref}}$ (95\% CI) \\
\midrule
Evaluator-specific & -0.011120 & 0.029466 & 0.040585 [0.030288, 0.049879] \\
Common $F_{\mathrm C}$ & -0.011120 & 0.009641 & 0.020760 [0.010131, 0.031402] \\
Common $F_{\mathrm S}$ & -0.018632 & 0.029466 & 0.048098 [0.037788, 0.057694] \\
\bottomrule
\end{tabular}

\end{table}

The evaluator-specific interaction is $+0.040585$ (95\% CI
$[+0.030288,+0.049879]$). The common controlled reference gives
$+0.020760$ ($[+0.010131,+0.031402]$). The common strong reference gives
$+0.048098$ ($[+0.037788,+0.057694]$). All three intervals exclude
zero. The common-reference contrasts therefore support an evaluator-associated
shift, while also showing that its magnitude depends on the reference protocol.

\begin{table}[H]
\centering
\caption{Evaluator-specific interactions across the two frozen policies,
five budgets, and three probabilistic metrics.}
\scriptsize
\begin{tabular}{llcr}
\toprule
Policy & Reference & $k$ & Metric: $I$ (95\% CI) \\
\midrule
ECG-on-Demand & Evaluator-specific & 2 & NLL: 0.0315 [0.0238, 0.0394] \\
ECG-on-Demand & Evaluator-specific & 2 & Brier: 0.0117 [0.0090, 0.0145] \\
ECG-on-Demand & Evaluator-specific & 2 & ECE: 0.0127 [0.0040, 0.0212] \\
ECG-on-Demand & Evaluator-specific & 3 & NLL: 0.0463 [0.0372, 0.0558] \\
ECG-on-Demand & Evaluator-specific & 3 & Brier: 0.0173 [0.0138, 0.0204] \\
ECG-on-Demand & Evaluator-specific & 3 & ECE: 0.0086 [0.0011, 0.0175] \\
ECG-on-Demand & Evaluator-specific & 4 & NLL: 0.0406 [0.0303, 0.0499] \\
ECG-on-Demand & Evaluator-specific & 4 & Brier: 0.0140 [0.0105, 0.0172] \\
ECG-on-Demand & Evaluator-specific & 4 & ECE: 0.0138 [0.0069, 0.0232] \\
ECG-on-Demand & Evaluator-specific & 6 & NLL: 0.0525 [0.0411, 0.0634] \\
ECG-on-Demand & Evaluator-specific & 6 & Brier: 0.0177 [0.0142, 0.0211] \\
ECG-on-Demand & Evaluator-specific & 6 & ECE: 0.0226 [0.0158, 0.0298] \\
ECG-on-Demand & Evaluator-specific & 8 & NLL: 0.0366 [0.0264, 0.0463] \\
ECG-on-Demand & Evaluator-specific & 8 & Brier: 0.0095 [0.0065, 0.0127] \\
ECG-on-Demand & Evaluator-specific & 8 & ECE: 0.0261 [0.0186, 0.0323] \\
MGA & Evaluator-specific & 2 & NLL: 0.0326 [0.0252, 0.0403] \\
MGA & Evaluator-specific & 2 & Brier: 0.0120 [0.0092, 0.0147] \\
MGA & Evaluator-specific & 2 & ECE: 0.0105 [0.0020, 0.0201] \\
MGA & Evaluator-specific & 3 & NLL: 0.0546 [0.0458, 0.0629] \\
MGA & Evaluator-specific & 3 & Brier: 0.0200 [0.0169, 0.0229] \\
MGA & Evaluator-specific & 3 & ECE: 0.0159 [0.0071, 0.0239] \\
MGA & Evaluator-specific & 4 & NLL: 0.0472 [0.0389, 0.0557] \\
MGA & Evaluator-specific & 4 & Brier: 0.0162 [0.0134, 0.0190] \\
MGA & Evaluator-specific & 4 & ECE: 0.0189 [0.0110, 0.0278] \\
MGA & Evaluator-specific & 6 & NLL: 0.0637 [0.0537, 0.0742] \\
MGA & Evaluator-specific & 6 & Brier: 0.0218 [0.0187, 0.0251] \\
MGA & Evaluator-specific & 6 & ECE: 0.0297 [0.0233, 0.0372] \\
MGA & Evaluator-specific & 8 & NLL: 0.0462 [0.0372, 0.0555] \\
MGA & Evaluator-specific & 8 & Brier: 0.0126 [0.0097, 0.0158] \\
MGA & Evaluator-specific & 8 & ECE: 0.0338 [0.0263, 0.0395] \\
\bottomrule
\end{tabular}

\end{table}

\section*{S12. Mask-distribution robustness}

The post-hoc mask-robust evaluator was trained with 50\% random
exact-cardinality masks, 25\% frozen ECG-on-Demand masks, and 25\% frozen MGA
masks.  The mask budgets were $\{1,2,3,4,6,8,12\}$.  Folds 1--6 supplied
training records, fold 7 supplied early stopping, and fold 10 was read only
for a boundary check; fold 8 was not used for training, tuning, or selection.
On fold 7 with all 12 channels, the mask-robust evaluator obtained NLL 0.2584,
Brier 0.0775, ECE 0.0263, AUROC 0.9345, and AUPRC 0.8425.  The corresponding
controlled-evaluator values were 0.5031, 0.1587, 0.1553, 0.8355, and 0.6283.

For the held-out fold-8 evaluation, $I_{\mathrm{robust}}$ is computed by
subtracting the controlled ECG-on-Demand contrast to $F_{\mathrm C}$ from
the mask-robust contrast to the same $F_{\mathrm C}$ reference.  At $k=4$,
the NLL interaction is $+0.037649$ (95\% CI
$[+0.026942,+0.047100]$), the Brier interaction is $+0.012817$
($[+0.009404,+0.016111]$), and the ECE interaction is $+0.018932$
($[+0.011015,+0.026595]$).  All 30 policy--budget--metric intervals exclude
zero.

\begin{table}[H]
\centering
\caption{Complete mask-robust evaluator interactions on fold 8.}
\scriptsize
\begin{tabular}{llcr}
\toprule
Policy & $k$ & Metric & $I_{\mathrm{robust}}$ (95\% CI) \\
\midrule
ECG-on-Demand & 2 & NLL & 0.0256 [0.0178, 0.0332] \\
ECG-on-Demand & 2 & Brier & 0.0097 [0.0071, 0.0125] \\
ECG-on-Demand & 2 & ECE & 0.0192 [0.0090, 0.0270] \\
ECG-on-Demand & 3 & NLL & 0.0384 [0.0289, 0.0479] \\
ECG-on-Demand & 3 & Brier & 0.0143 [0.0110, 0.0174] \\
ECG-on-Demand & 3 & ECE & 0.0158 [0.0075, 0.0232] \\
ECG-on-Demand & 4 & NLL & 0.0376 [0.0269, 0.0471] \\
ECG-on-Demand & 4 & Brier & 0.0128 [0.0094, 0.0161] \\
ECG-on-Demand & 4 & ECE & 0.0189 [0.0110, 0.0266] \\
ECG-on-Demand & 6 & NLL & 0.0511 [0.0400, 0.0624] \\
ECG-on-Demand & 6 & Brier & 0.0174 [0.0141, 0.0209] \\
ECG-on-Demand & 6 & ECE & 0.0241 [0.0167, 0.0300] \\
ECG-on-Demand & 8 & NLL & 0.0312 [0.0212, 0.0414] \\
ECG-on-Demand & 8 & Brier & 0.0078 [0.0047, 0.0110] \\
ECG-on-Demand & 8 & ECE & 0.0202 [0.0145, 0.0269] \\
MGA & 2 & NLL & 0.0258 [0.0181, 0.0333] \\
MGA & 2 & Brier & 0.0098 [0.0072, 0.0123] \\
MGA & 2 & ECE & 0.0156 [0.0066, 0.0249] \\
MGA & 3 & NLL & 0.0455 [0.0363, 0.0543] \\
MGA & 3 & Brier & 0.0166 [0.0136, 0.0195] \\
MGA & 3 & ECE & 0.0195 [0.0118, 0.0268] \\
MGA & 4 & NLL & 0.0437 [0.0354, 0.0519] \\
MGA & 4 & Brier & 0.0152 [0.0123, 0.0181] \\
MGA & 4 & ECE & 0.0210 [0.0132, 0.0285] \\
MGA & 6 & NLL & 0.0653 [0.0555, 0.0747] \\
MGA & 6 & Brier & 0.0217 [0.0186, 0.0248] \\
MGA & 6 & ECE & 0.0323 [0.0260, 0.0390] \\
MGA & 8 & NLL & 0.0448 [0.0348, 0.0546] \\
MGA & 8 & Brier & 0.0121 [0.0092, 0.0152] \\
MGA & 8 & ECE & 0.0322 [0.0257, 0.0387] \\
\bottomrule
\end{tabular}

\end{table}

\section*{S13. Evaluator-aligned Strong-MGA}

Strong-MGA uses the original strong evaluator to generate one-step marginal
NLL-reduction targets from training records only.  It retains the MGA policy
architecture and optimization settings, but changes the target evaluator.
No fold-8 state or label was used during target generation or fitting.

\begin{table}[H]
\centering
\caption{Strong-evaluator performance at $k=4$ for the frozen controlled-
trained policies and post-hoc Strong-MGA.  Delta is adaptive minus fixed
$F_{\mathrm S}$; negative values favor adaptive allocation.}
\scriptsize
\begin{tabular}{lrrrr}
\toprule
Method & Metric & Adaptive & Fixed $F_{\mathrm S}$ & Delta (95\% CI) \\
\midrule
ECG-on-Demand (C-trained) & NLL & 0.2955 & 0.2660 & 0.0295 [0.0228, 0.0368] \\
ECG-on-Demand (C-trained) & Brier & 0.0907 & 0.0817 & 0.0090 [0.0065, 0.0114] \\
ECG-on-Demand (C-trained) & ECE & 0.0165 & 0.0244 & -0.0079 [-0.0141, 0.0005] \\
MGA (C-trained) & NLL & 0.2952 & 0.2660 & 0.0291 [0.0223, 0.0361] \\
MGA (C-trained) & Brier & 0.0906 & 0.0817 & 0.0088 [0.0065, 0.0111] \\
MGA (C-trained) & ECE & 0.0181 & 0.0244 & -0.0063 [-0.0132, 0.0026] \\
Strong-MGA (S-target) & NLL & 0.2784 & 0.2660 & 0.0124 [0.0058, 0.0192] \\
Strong-MGA (S-target) & Brier & 0.0855 & 0.0817 & 0.0037 [0.0014, 0.0062] \\
Strong-MGA (S-target) & ECE & 0.0203 & 0.0244 & -0.0041 [-0.0114, 0.0029] \\
\bottomrule
\end{tabular}

\end{table}

Strong-MGA reduces NLL relative to controlled-trained MGA by $-0.016777$
(95\% CI $[-0.025963,-0.008355]$) and Brier by $-0.005129$
($[-0.007872,-0.002205]$).  Its strong-evaluator NLL gap to exhaustive
$F_{\mathrm S}$ remains positive at $+0.012362$ ($[+0.005769,+0.019204]$),
so alignment gives partial recovery rather than complete recovery.

\begin{table}[H]
\centering
\caption{Paired Strong-MGA versus controlled-trained MGA differences at $k=4$.
Negative values favor Strong-MGA.}
\small
\begin{tabular}{lrr}
\toprule
Metric & Strong-MGA minus MGA & 95\% CI \\
\midrule
NLL & -0.0168 & [-0.0260, -0.0084] \\
Brier & -0.0051 & [-0.0079, -0.0022] \\
ECE & 0.0021 & [-0.0045, 0.0066] \\
\bottomrule
\end{tabular}

\end{table}

\section*{S14. Evidence boundary, limitations, and provenance}

The original frozen-policy analysis supports a controlled-evaluator advantage
and a positive evaluator interaction on the held-out fold 8.  The additions in
S11--S13 were motivated after that result: they are robustness and mechanism
analyses, not prespecified confirmatory extensions.  Their shared patient
roles, frozen references, bootstrap seed, and no-fold-8 training boundary
make them auditable, but post-hoc motivation limits the strength of their
inferential interpretation.

Fold 8 appeared in historical pre-reset training pools.  All current models
were retrained without fold 8, and fold 8 was excluded from training, early
stopping, fixed-subset selection, Strong-MGA target generation, and policy
selection for the revision analyses.  We therefore report it as a held-out
exploratory evaluation rather than an untouched confirmatory test.  A
post-run provenance review also found that a manually entered display
timestamp did not match filesystem chronology and that the initial start
receipt was not preserved as a separate immutable file; neither issue changed
data access, predictions, resamples, metrics, models, or conclusions.

The retrospective study is restricted to PTB-XL and masks simultaneously
recorded channels rather than modeling prospective electrode placement,
workflow, safety, fairness, or clinical replacement of a 12-lead recording.
The fold-10 label-aware oracle is non-deployable and evaluator/loss-specific.
No equivalence or noninferiority margin was defined, and no prospective
clinical effectiveness claim is made.  Strong-MGA is a one-step aligned policy,
not a jointly optimized end-to-end sensing--diagnosis system.

\section*{S15. Selection-set channel heterogeneity}
Main-paper Fig.~1(c--e) reports the channel sets, oracle first-lead choices,
and fixed-versus-oracle NLL headroom. The fold-10 oracle quantities are
label-aware, controlled-evaluator-specific, descriptive, and non-deployable.
The modal policy sets use the development fold identified in the figure.
These panels are not repeated here. Complete fold-8 controlled-evaluator
contrasts are provided in S9.